\PassOptionsToPackage{table}{xcolor}
\documentclass[sigconf,natbib=true,nonacm,pbalance]{acmart}
\AtBeginDocument{%
  }

\usepackage[frozencache,cachedir=minted]{minted}
\setminted{style=trac}
\newminted{sparql}{escapeinside=``}
\newmintinline{sparql}{escapeinside=``}
\usepackage{etoolbox}
\robustify{\sparqlinline}
\usepackage{multirow}
\usepackage{multicol}
\usepackage{threeparttable}
\usepackage{calc}
\usepackage{enumitem}
\usepackage{cleveref}
\newcommand{\github}{\href{https://github.com/ad-freiburg/wikidata-query-logs}{github.com/ad-freiburg/wikidata-query-logs}}
\newcommand{\wdql}{\href{https://wdql.cs.uni-freiburg.de}{wdql.cs.uni-freiburg.de}}
\newcommand{\grasp}{\href{https://github.com/ad-freiburg/grasp}{github.com/ad-freiburg/grasp}}
\newcommand{\graspeval}{\href{https://grasp.cs.uni-freiburg.de/evaluate}{grasp.cs.uni-freiburg.de/evaluate}}

\newcommand{\grisp}{\href{https://grasp.cs.uni-freiburg.de/grisp}{\nolinkurl{grasp.cs.uni-freiburg.de/grisp}}}
\newcommand{\grispdata}{\href{https://ad-publications.cs.uni-freiburg.de/grisp}{\nolinkurl{ad-publications.cs.uni-freiburg.de/grisp}}}

\definecolor{fncolor}{RGB}{0, 1, 73}
\def\FN#1{{\color{fncolor} \texttt{#1}}}
\definecolor{fnsetcolor}{RGB}{137, 83, 148}

\begin{document}

\title{The Wikidata Query Logs Dataset}

\author{Sebastian Walter}
\email{swalter@cs.uni-freiburg.de}
\affiliation{%
  \institution{University of Freiburg}
  \department{Department of Computer Science}
  \city{Freiburg im Breisgau}
  \country{Germany}
}

\author{Hannah Bast}
\email{bast@cs.uni-freiburg.de}
\affiliation{%
	\institution{University of Freiburg}
	\department{Department of Computer Science}
	\city{Freiburg im Breisgau}
	\country{Germany}
}

\begin{abstract}
We present the \textbf{Wikidata Query Logs (WDQL) dataset}, a dataset consisting of \textbf{335k question-query pairs} over the Wikidata knowledge graph. It is \textbf{over 11x larger} than the largest existing Wikidata datasets of similar format without relying on template-generated queries. Instead, we construct it using real-world SPARQL queries sent to the Wikidata Query Service and generate questions for them. Since these log-based queries are anonymized, and therefore often do not produce results, a significant amount of effort is needed to convert them back into meaningful SPARQL queries. To achieve this, we present an agent-based method that iteratively de-anonymizes, cleans, and verifies queries against Wikidata while also generating corresponding natural-language questions. We demonstrate the benefit of this dataset for training question-answering methods. All WDQL assets, as well as the agent code, are publicly available via \github{} under a permissive license.
\end{abstract}


\maketitle

\section{Introduction}

Wikidata \cite{wikidata} is the currently largest open general-purpose knowledge graph. At the time of this writing, the full RDF
dump contains around 20 billion triples describing over 120 million items.
The standard way to query Wikidata is via the Wikidata Query Service (WDQS), a public SPARQL endpoint that has been in operation since September 7, 2015; see \href{https://query.wikidata.org}{query.wikidata.org}.

SPARQL is a powerful query language for RDF data, allowing users to express complex queries with precise semantics. However, SPARQL can be difficult to learn and use effectively, even for experts. For many purposes, it would be desirable to be able to specify a question in natural language and have the system translate it into a SPARQL query automatically.
Thanks to the spectacular advances in deep learning, several systems now exist for this purpose. Most of these systems rely on training sets consisting of pairs of natural-language questions and SPARQL queries. Indeed, several such datasets have been published in recent years; see \Cref{sec:related} for an overview.

In this paper, we present the largest dataset of this kind to date.\footnote{WDQL is a continuously evolving dataset. Based on the current state of the processing pipeline, we estimate a potential of up to 730k final pairs. See \github{} for more information.} It automatically derives question-query pairs from the WDQS query logs, despite IRIs and literals in the queries being anonymized. More generally, our methothology can be used to turn any given SPARQL query log
(even when anonymized) into a dataset of this kind.

\subsection*{Core contributions}
1. We provide a dataset of 335k question-query pairs for Wikidata, derived from real-world SPARQL queries sent to the Wikidata Query Service. We call it the Wikidata Query Logs (WDQL) dataset.\\[1mm]
2. We demonstrate the dataset's effectiveness for the downstream task of question answering over Wikidata.\\[1mm]
3. We provide all dataset assets and code for download and reproduction openly at \github{} and \grasp{}. This includes a visualization app available at \wdql.\\[1mm]
4. More generally, our methodology (and code) can be used to derive question-query pairs from other query logs as well.

\begin{table*}
	\centering
	\caption{Summary statistics for WDQL and other Wikidata datasets}
	\begin{threeparttable}
\begin{tabular}{l@{\hskip 0.25cm}rrrrrr}
\toprule
\textbf{Dataset} & \textbf{Queries} & \textbf{Distinct IRIs} & \textbf{Avg. Triples} & \textbf{Advanced (\%)\tnote{1}} & \textbf{SELECT (\%)} & \textbf{ASK (\%)} \\
\midrule
SimpleQuestions & 27,924 & 24,485 & 1.00 & 0.0 & 100.0 & -- \\
LC-QuAD 2.0 & 30,152 & 27,223 & 2.07 & 0.0 & 91.1 & 8.9 \\
WWQ & 4,316 & 2,581 & 1.24 & 13.9 & 100.0 & -- \\
QAWiki & 518 & 810 & 2.38 & 60.6 & 93.6 & 6.8 \\
QALD-7 & 150 & 283 & 1.69 & 23.3 & 92.7 & 8.0 \\
QALD-10 & 806 & 1,107 & 1.51 & 27.9 & 92.9 & 7.6 \\
SPINACH & 320 & 876 & 4.62 & 87.0 & 100.0 & -- \\
\midrule
WDQL & 335,450 & 85,350 & 3.18 & 55.5 & 94.9 & 5.1 \\
\end{tabular}
\end{threeparttable}
\vspace{0.1cm}
\par
\centering
\begin{minipage}[t]{\widthof{\footnotesize\textsuperscript{1}~SPARQL queries using constructs beyond basic features (triple patterns, FILTER, ORDER BY, LIMIT, OFFSET, DISTINCT, GROUP BY and aggregate functions)}}\vspace{0pt}
\centering
\footnotesize\raggedright
\textsuperscript{1}~SPARQL queries using constructs beyond basic features (triple patterns, FILTER, ORDER BY, LIMIT, OFFSET, DISTINCT, GROUP BY and aggregate functions)
\end{minipage}

	\label{tab:stats}
\end{table*}

\section{Related Work}\label{sec:related}

A number of related datasets providing paired data of questions and SPARQL queries exist for the Wikidata knowledge graph, which we will briefly outline and contrast to WDQL in the following.\\[1mm]
\textbf{SimpleQuestions \cite{simplequestions}} is a dataset consisting of simple, single-triple SPARQL queries and corresponding human-annotated questions. It is a Wikidata port of \cite{simplequestionsfb} which was originally built using facts from a subset of the Freebase \cite{freebase} knowledge graph.\\[1mm]
\textbf{QALD-7 \cite{qald7} and QALD-10 \cite{qald10}} are the first and latest datasets in the QALD series, respectively, that provide pairs of questions and SPARQL queries over Wikidata. Both are built from real-world, manually created questions which were annotated with corresponding SPARQL queries by human experts.\\[1mm]
\textbf{LC-QuAD 2.0 \cite{lcquad2}} generates SPARQL queries and corresponding template questions using a pre-defined set of SPARQL query templates. The template questions are then verbalized into natural-language questions and paraphrased via crowd sourcing.\\[1mm]
\textbf{WikiWebQuestions (WWQ) \cite{wikisp}} is a Wikidata port of the popular WebQuestionsSP dataset \cite{wqsp}, that annotates questions derived from Google's Suggest API with SPARQL queries over Freebase.\\[1mm]
\textbf{SPINACH \cite{spinach}} is a dataset created from questions and SPARQL queries on Wikidata's ``Request a query'' forum\footnote{See \href{https://www.wikidata.org/wiki/Wikidata:Request_a_query}{www.wikidata.org/wiki/Wikidata:Request\_a\_query}}, where users can ask other users for help with writing SPARQL queries.\\[1mm]
\textbf{QAWiki \cite{qawiki}} is a collaborative repository for sharing questions and corresponding SPARQL queries over various knowledge graphs, but is mainly used for Wikidata at the moment. We use QAWiki v1, a snapshot from Sept. 2025 with 518 question-query pairs.\\[1mm]
All datasets differ from WDQL in multiple ways: QALD-7, QALD-10, QAWiki, and SPINACH are built from human-derived questions and SPARQL queries, which provide high quality but on a small scale. WWQ contains a few thousand questions, though most can be answered with relatively simple SPARQL queries (see \Cref{tab:stats}). SimpleQuestions is larger, but restricted to single-triple SPARQL queries. LC-QuAD 2.0 achieves larger scale through template-generated SPARQL queries, which limit diversity to specific query patterns. In contrast, WDQL is built from real-world SPARQL queries submitted to WDQS and classified to be of human origin, for which we generate questions using an agent based on a large language model (LLM). This approach allows us to achieve a 6x larger scale than previous datasets while maintaining high diversity and real-world grounding through actual user queries.

\section{Dataset}

In this section we present the WDQL dataset. We start with the underlying agent for generating questions from SPARQL queries, followed by describing the dataset creation process. We then compare WDQL with other Wikidata datasets.

\subsection{SPARQL-to-Question Agent}

We base our agent design and implementation on the GRASP method and system \cite{grasp,graspdemo}. GRASP is an agentic method originally developed for SPARQL-based question answering (SPARQL QA), that works by interactively exploring a knowledge graph via function calls. In prior work, GRASP was extended to other tasks, for example for linking table data to knowledge graph entities \cite{graspel}.

We follow this approach and implement our SPARQL-to-Question (S2Q) agent as a new GRASP task. We keep the core functions of GRASP for knowledge graph interactions, which are \FN{EXE} (execute SPARQL query), \FN{LST} (list triples given constraints), \FN{SEN} (search for entity), \FN{SPR} (search for property), \FN{SPE} (search for property of a given entity), \FN{SOP} (search for object of a given property), \FN{SCN} (search for items given triple constraints), and \FN{SAC} (search for items given a constraining SPARQL query). We then adapt the \FN{ANS} (answer and stop) and \FN{CAN} (cancel and stop) from SPARQL QA to S2Q as follows:\\[1mm]
\FN{ANS: answer(questions: list[str], sparql: str)}\\
Provide a list of questions capturing the intent of the SPARQL query, as well as a cleaned version of the input SPARQL query, and stop.\\[1mm]
\FN{CAN: cancel(reason: str)}\\
Provide a reason why questions for the input SPARQL query cannot be generated and stop.\\[1mm]
We also adapt the system instructions of the GRASP agent: First, it should think about the intent behind the input SPARQL query, and retrieve additional context from the knowledge graph via the available functions if it cannot be determined from the SPARQL query alone. Then, it should ``clean'' the SPARQL query, which includes: assigning proper names to variables, removing superfluous SPARQL constructs, or replacing anonymized string literals with sensible values\footnote{This part is specifically tailored for the anonymized nature of Wikidata SPARQL Logs, see \Cref{subsec:creation}.}. In this stage, the agent is allowed to deviate from the original input SPARQL structure if it would make the resulting SPARQL query execute properly, more natural, or more precise. The agent is also instructed to execute the final cleaned SPARQL query to make sure it returns the intended results, and reiterate if it does not. Finally, we ask the agent to generate one, two, or three natural-language questions for the cleaned SPARQL query with different phrasings and levels of detail before calling the \FN{ANS} function. At any point, the agent can call \FN{CAN} to stop the process. For example, if it cannot make the input SPARQL query work without deviating too much from its original intent.

Note that because GRASP is designed to interact with arbitrary knowledge graphs, this agent can be applied to logs of SPARQL queries for other knowledge graphs as well.

\subsection{Creation}\label{subsec:creation}

\begin{table*}
	\centering
	\caption{Percentage of queries using specific SPARQL constructs for WDQL and other Wikidata datasets }
	\begin{tabular}{l@{\hskip 0.15cm}cccccccccccc}
\toprule
\textbf{Dataset} & \textbf{Filt} & \textbf{Ord} & \textbf{Lim} & \textbf{Grp} & \textbf{Agg} & \textbf{Opt} & \textbf{Un} & \textbf{Min} & \textbf{Val} & \textbf{Func} & \textbf{Paths} & \textbf{Subq} \\
\midrule
SimpleQuestions & -- & -- & -- & -- & -- & -- & -- & -- & -- & -- & -- & -- \\
LC-QuAD 2.0 & 24.5 & 3.8 & 12.4 & -- & 4.7 & -- & -- & -- & -- & 18.9 & -- & -- \\
WWQ & 3.6 & 2.9 & 3.5 & \textless0.1 & 0.1 & -- & 0.3 & -- & -- & \textless0.1 & 13.0 & -- \\
QAWiki & 10.8 & 15.1 & 15.1 & 4.4 & 10.0 & 1.4 & 0.6 & 6.2 & 0.4 & 9.1 & 56.6 & 3.3 \\
QALD-7 & 7.3 & 7.3 & 7.3 & 3.3 & 6.7 & -- & 6.7 & -- & -- & 2.0 & 19.3 & 0.7 \\
QALD-10 & 12.7 & 8.2 & 8.2 & 1.5 & 16.9 & 0.1 & 2.0 & 0.4 & 0.6 & 9.4 & 20.8 & 1.4 \\
SPINACH & 37.5 & 28.7 & 4.4 & 18.1 & 15.7 & 36.2 & 8.9 & 8.5 & 12.6 & 30.7 & 43.0 & 3.4 \\
\midrule
WDQL & 62.5 & 19.8 & 26.7 & 11.5 & 12.3 & 28.6 & 5.1 & 1.6 & 7.3 & 59.1 & 16.6 & 2.5 \\
\end{tabular}
\vspace{0.1cm}
\par
\centering
\begin{minipage}[t]{\textwidth}\vspace{0pt}
\footnotesize\raggedright
\textbf{Filt} = FILTER; \textbf{Ord} = ORDER BY; \textbf{Lim} = LIMIT/OFFSET; \textbf{Grp} = GROUP BY; \textbf{Agg} = Aggregate functions (COUNT, SUM, GROUP\_CONCAT, etc.); \textbf{Opt} = OPTIONAL; \textbf{Un} = UNION; \textbf{Min} = MINUS; \textbf{Val} = VALUES; \textbf{Func} = Non-aggregate functions (LANG, REGEX, IF, etc., including BIND); \textbf{Paths} = Property paths and modifiers (|, /, ?, *, +); \textbf{Subq} = Subqueries
\end{minipage}

	\label{tab:sparql}
\end{table*}

We start by downloading the organic split of the Wikidata SPARQL Logs \cite{querylogs}, which contain SPARQL queries classified as sent by a human user to WDQS. The logs are provided in seven intervals between June 2017 and March 2018 and contain a total of around 3.5 million SPARQL queries. Importantly, the SPARQL queries are anonymized, which means that comments are removed, string literals longer than 10 characters are normalized to \sparqlinline|"string1"|, \sparqlinline|"string2"|, etc., variables are normalized to \sparqlinline|?v1|, \sparqlinline|?v2|, etc., and more. See the accompanying \href{https://iccl.inf.tu-dresden.de/web/Wissensbasierte_Systeme/WikidataSPARQL}{website} for full details about the anonymization. For our goal of generating questions from SPARQL queries, the anonymization is an additional hurdle to overcome, because it drops valuable information from comments, variable names, and literals about the query intent and also often causes the SPARQL query to produce empty results. We deduplicate the organic SPARQL logs and end up with around 860k distinct queries.

To make the task for our S2Q agent a bit easier, we do not only input the raw anonymized SPARQL query, but pre-process and extend it as follows:\\[1mm]
1. We remove all occurrences of the \sparqlinline|wikibase:label| service, which appears in many of the queries, but is not compliant with the SPARQL standard. We add an instruction for the agent to explicitly add labels via \sparqlinline|rdfs:label| instead, where appropriate. We also remove all unused variables in the query's \sparqlinline|SELECT| clause.\\[1mm]
2. For all IRIs in the SPARQL queries, we also add their labels, aliases, and other additional information, such as \sparqlinline|schema:description|.\\[1mm]
3. We execute the SPARQL query and add its execution result formatted as a markdown table. For all IRIs in the execution result, we again also specify their labels to make the result interpretable without requiring the agent to look them up individually.\\[1mm]
We ran the agent with Qwen3 Next 80B A3B Instruct \cite{qwen3} as underlying model, first on two NVIDIA H200 GPUs in BF16 precision and later on one NVIDIA H200 GPU in FP8 precision, then switched to Qwen3.5 27B \cite{qwen35} on one NVIDIA H100 GPU in BF16 precision. For the SPARQL endpoint we use QLever \cite{qlever} and a recent Wikidata dump. On average, the agent takes \textasciitilde{}30s to clean and generate questions for one SPARQL query. After about five months (with some pauses and downtime due to fluctuating GPU availability) we processed \textasciitilde{}450k SPARQL queries. The agent produced an invalid output (e.g., due to function call failures) for \textasciitilde{}86k SPARQLs and explicitly cancelled the processing of \textasciitilde{}4k queries via \FN{CAN}, leaving us with \textasciitilde{}360k cleaned SPARQL queries for which it generated corresponding questions. From these \textasciitilde{}360k queries we further remove all instances where the SPARQL query cannot be parsed, or executed (e.g., because they produce OOM errors), or has an empty result. This leaves us with a final set of 335,450 valid question-query pairs. See the GitHub repository at \github{} for more detailed numbers.

Next, we compute embeddings of all questions in the final dataset with \href{https://huggingface.co/Qwen/Qwen3-Embedding-0.6B}{Qwen3 Embedding 0.6B}, and we use the average embedding of all questions per SPARQL query as input to UMAP \cite{umap} to generate both 2-dimensional and 50-dimensional representations of all question-query pairs. The 2-dimensional representations are used for visualization, the 50-dimensional representations for clustering. We cluster the pairs using HDBSCAN \cite{hdbscan} with liberal parameters in order to find near-duplicate pairs. For example, the largest cluster contains 165 pairs where the questions are of the form ``What movie has Sratim ID ...?''.  All pairs which are labeled as outliers by HDBSCAN are kept and assigned to their own cluster. We then use the clusters to determine a split of our dataset for training knowledge graph question-answering (KGQA) methods. We split the clusters into 80\% for training, and 10\% for both validation and test. All pairs in each cluster are then assigned to the corresponding split, and we end up with \textasciitilde{}268.9k train, \textasciitilde{}33.1k validation, and \textasciitilde{}33.4k test pairs.

\begin{table*}
	\caption{Prefix, query pattern, and literal statistics for WDQL and other Wikidata datasets}
	\begin{threeparttable}
\begin{tabular}{l@{\hskip 0.15cm}rrrrrrrrrr}
\toprule
\multirow{2}{*}{\textbf{Dataset}} & \multicolumn{6}{c}{\textbf{Prefixes}\tnote{1}} & \multicolumn{2}{c}{\textbf{Patterns}\tnote{2}} & \multicolumn{2}{c}{\textbf{Other}\tnote{3}} \\
\cmidrule(lr){2-7}\cmidrule(lr){8-9}\cmidrule(lr){10-11}
& Ent+Dir & Stmt+Qual & Ref & Reif & Adv & Cov & \# Dist & \% Uniq & Lit & Lang \\
\midrule
SimpleQuestions & 24,485 & -- & -- & -- & -- & 2/19 & 2 & 0.0 & 0 & 0 \\
LC-QuAD 2.0 & 26,336 & 887 & -- & -- & -- & 5/19 & 45 & 0.1 & 3,689 & 1 \\
WWQ & 2,483 & 95 & -- & -- & 3 & 6/19 & 187 & 4.3 & 52 & 1 \\
QAWiki & 683 & 101 & -- & -- & 26 & 8/19 & 334 & 64.5 & 21 & 0 \\
QALD-7 & 265 & 18 & -- & -- & -- & 5/19 & 68 & 45.3 & 9 & 1 \\
QALD-10 & 1,040 & 57 & -- & -- & 10 & 7/19 & 352 & 43.7 & 49 & 4 \\
SPINACH & 634 & 190 & 6 & -- & 46 & 10/19 & 293 & 100.0 & 170 & 7 \\
\midrule
WDQL & 80,730 & 2,257 & 345 & 1,759 & 259 & 16/19 & 125,892 & 37.5 & 83,798 & 184 \\
\end{tabular}
\end{threeparttable}
\vspace{0.1cm}
\par
\centering
\begin{minipage}[t]{\textwidth}\vspace{0pt}
\footnotesize\raggedright
\textsuperscript{1}~Number of distinct IRIs by Wikidata prefix. Ent+Dir = \sparqlinline|wd:| + \sparqlinline|wdt:| (entities and direct properties), Stmt+Qual = \sparqlinline|p:| + \sparqlinline|ps:| + \sparqlinline|pq:| (statements and qualifiers), Ref = \sparqlinline|pr:| + \sparqlinline|prv:| + \sparqlinline|wdref:| (references), Reif = \sparqlinline|wds:| + \sparqlinline|wdv:| (reification nodes), Adv = \sparqlinline|wdtn:| + \sparqlinline|psn:| + \sparqlinline|pqn:| + \sparqlinline|psv:| + \sparqlinline|pqv:| + \sparqlinline|wdno:| + \sparqlinline|wikibase:| (advanced prefixes), Cov\ = Number of prefixes covered out of all 19 \href{https://www.mediawiki.org/wiki/Wikibase/Indexing/RDF_Dump_Format#Prefixes_used}{prefixes supported by the Wikidata Query Service}; \sparqlinline|prn:|, \sparqlinline|pqn:|, and \sparqlinline|wdata:| do not occur in any dataset\\
\textsuperscript{2}~\# Dist = Number of distinct query patterns after normalizing variables, entities, properties, and literals; \% Uniq = Share of queries with a unique query pattern (\# Dist / \# Total)\\
\textsuperscript{3}~Lit\ = Number of distinct literals; Lang\ = Number of distinct valid (BCP 47) languages used in language-related filter expressions
\end{minipage}

	\label{tab:ext}
\end{table*}

\subsection{Statistics and Comparison}

We calculate the statistics for all following tables across all SPARQL queries from train, validation, and test splits that can be parsed according to SPARQL 1.1. This is the case for all queries for all datasets except for SPINACH (293/320) and LC-QuAD 2.0 (30,145/30,152).

An overview of WDQL and other Wikidata datasets is shown in \Cref{tab:stats}. The standout property of WDQL is its size at over 335k queries. Compared to the previous largest Wikidata dataset, LC-QuAD 2.0, WDQL is over 11x larger and at the same time contains queries with more triple patterns and advanced SPARQL constructs on average. \Cref{tab:sparql} provides more details about the SPARQL constructs present in queries of the different datasets. WDQL and SPINACH are the only datasets that cover all of the most commonly used SPARQL constructs to a significant extent.

When we look at the distribution of IRIs by prefix in each dataset, we find that WDQL has the broadest coverage with 16 out of 19 prefixes, and is also the only dataset that has queries using Wikidata's reification predicates. Moreover, we find that 37.5\% (126k) of WDQL's queries have a unique pattern, that is, no other query is the same after normalizing variables, entities, properties, and literals. This is a lower percentage of queries than for QAWiki, QALD-7, QALD-10, and SPINACH (in which all queries have a unique pattern), but orders of magnitude higher than for SimpleQuestions, LC-QuAD 2.0, and WWQ. Given the size of WDQL, 37.5\% is higher than we would have expected. Finally, WDQL contains many more distinct literals and languages used in language filters than the other datasets. See the detailed statistics in \Cref{tab:ext}.

We also compare the SPARQL queries from the final question-query pairs against the raw SPARQL queries from WDQS to assess how close our S2Q agent stays to the original queries. Most notably, our agent slightly reduces the average number of triples ($3.58 \to 3.18$), as well as the share of SPARQL queries with sub-queries ($7.3\% \to 2.5\%$) and \sparqlinline|UNION| constructs ($9.1\% \to 5.1\%$), but increases the share of SPARQL queries with a unique query pattern ($28.7\% \to 37.5\%$). Additionally, we compute two Jaccard similarities for each pair of final and raw SPARQL queries, the first between their sets of IRIs, and the second between their sets of SPARQL constructs. For computing these similarities, we ignore \sparqlinline|SERVICE wikibase:label| in the raw SPARQL query, and triples using \sparqlinline|rdfs:label| and \sparqlinline|FILTER(LANG(?var) = "..")| in the final SPARQL query, because the WDQS label service is almost always replaced by \sparqlinline|rdfs:label| with an explicit language filter in our pipeline. We find that the mean similarity over all pairs is high for both types of sets, with 0.75 (median 1.0, 51.7\% perfect overlap, 2.8\% no overlap) and 0.75 (median 0.83, 46.9\% perfect overlap, 0.5\% no overlap) for IRIs and SPARQL constructs respectively. Overall, the results suggest that our agent tends to stay close to the raw SPARQL query, with occasional deviations that decrease complexity, but increase diversity. Note that these findings might change when using other base LLMs with our agent.

\begin{table*}
	\centering
	\caption{KGQA performance of GRISP for different training dataset compositions. Left: F\textsubscript{1}-score when trained on previous Wikidata datasets, WDQL only, or both. Right: LLM-as-a-judge evaluation with OpenAI GPT-4.1 mini between GRISP\textsubscript{Prev} and GRISP\textsubscript{WDQL}, showing preference and tie percentages. The evaluation reveals that SimpleQuestions, LC-QuAD 2.0, and WWQ should not be used to assess natural generalization capabilities of KGQA methods.}
	\label{tab:qa-combined}
	\begin{minipage}[t]{0.5\textwidth}\vspace{0pt}
\centering
\begin{tabular}{lccc}
\toprule
\textbf{Dataset} & \textbf{GRISP\textsubscript{Prev}} & \textbf{GRISP\textsubscript{WDQL}} & \textbf{GRISP\textsubscript{All}} \\
\midrule
SimpleQ & \cellcolor{gray!15}\textbf{90.5} & 41.4 & \cellcolor{gray!15}\underline{90.4} \\
LC-QuAD 2.0 & \cellcolor{gray!15}\textbf{73.3} & 42.6 & \cellcolor{gray!15}\underline{70.6} \\
WWQ & \cellcolor{gray!15}\textbf{78.2} & 54.5 & \cellcolor{gray!15}\underline{78.1} \\
QALD-7 & \cellcolor{gray!15}61.6 & \textbf{70.1} & \cellcolor{gray!15}\underline{63.4} \\
QALD-10 & \cellcolor{gray!15}32.4 & \textbf{49.9} & \cellcolor{gray!15}\underline{40.3} \\
QAWiki\textsuperscript{1} & 32.6 & \textbf{47.3} & \underline{45.0} \\
SPINACH\textsuperscript{1} & 4.1 & \underline{29.4} & \textbf{32.5} \\
WDQL\textsuperscript{2} & 6.8 & \cellcolor{gray!15}\textbf{57.6} & \cellcolor{gray!15}\underline{56.9} \\
\end{tabular}
\end{minipage}%
\hfill%
\begin{minipage}[t]{0.5\textwidth}\vspace{0pt}
\centering
\begin{tabular}{lccc}
\toprule
\textbf{Dataset} & \textbf{GRISP\textsubscript{Prev}} & \textbf{GRISP\textsubscript{WDQL}} & \textbf{Tie} \\
\midrule
SimpleQ & \cellcolor{gray!15}\underline{28.1\%} & \textbf{46.9\%} & 25.0\% \\
LC-QuAD 2.0 & \cellcolor{gray!15}30.9\% & \textbf{36.1\%} & \underline{33.0\%} \\
WWQ & \cellcolor{gray!15}\underline{33.1\%} & \textbf{51.1\%} & 15.8\% \\
QALD-7 & \cellcolor{gray!15}\underline{20.0\%} & \textbf{62.0\%} & 18.0\% \\
QALD-10 & \cellcolor{gray!15}15.7\% & \textbf{55.6\%} & \underline{28.7\%} \\
QAWiki\textsuperscript{1} & 16.8\% & \textbf{59.8\%} & \underline{23.4\%} \\
SPINACH\textsuperscript{1} & 3.0\% & \textbf{72.1\%} & \underline{24.8\%} \\
WDQL\textsuperscript{2} & 4.2\% & \cellcolor{gray!15}\textbf{87.7\%} & \underline{8.1\%} \\
\end{tabular}
\end{minipage}
\vspace{0.1cm}
\par
\centering
\begin{minipage}[t]{\textwidth}\vspace{0pt}
\footnotesize\raggedright
Bold and underlined values indicate the highest and second-highest scores for each dataset, respectively. Grey cells represent in-distribution test sets with samples from the corresponding training set in the training data. We adjusted the number of training epochs for GRISP\textsubscript{Prev} to 7.4 and GRISP\textsubscript{All} to 2.6 in order to match the number of samples seen during training of GRISP\textsubscript{WDQL}, which is trained for 4 epochs.\\
\textsuperscript{1}~SPINACH and QAWiki do not have a dedicated training set, and are considered out-of-distribution for all GRISP models.\\
\textsuperscript{2}~Evaluated on 1000 randomly selected test samples due to the large test set size.
\end{minipage}

\end{table*}

\section{Case Study: Question Answering}

In this section, we demonstrate that KGQA models achieve substantial improvements in downstream performance and generalization capabilities when trained on WDQL. All experiments use the initial WDQL release from February 2, 2026 with 200,186 question-query pairs. To save training time, we create a second KGQA export of this release, where we again split by cluster, but only retain one random query from each cluster instead of all. The resulting dataset has 103,327 pairs, about half the size of the full release, with 82,661 for training, 10,333 for validation, and 10,333 for testing. It is used to train and evaluate all models in the following. To ensure a fair evaluation and no leakage from WDQL into other datasets, we verify that no SPARQL query from our training set occurs verbatim in any test set. Specifically, we compute for each of the other benchmarks, the overlap (regarding query patterns, see above) of the WDQL training set with that benchmark's test set, and the internal overlap of the benchmark's own training set with its test set. We find that the WDQL overlap is never significantly higher, and often much lower than the benchmark's internal overlap (e.g., for WWQ and LC-QuAD~2.0). All experiments are based on GRISP \cite{grisp}, a recent SPARQL-based KGQA method achieving state-of-the-art on Wikidata in the fine-tuning setting, with Qwen2.5 7B as base model.

\Cref{tab:qa-combined} shows the results when training GRISP on the existing Wikidata datasets as GRISP\textsubscript{Prev}, only on our dataset as GRISP\textsubscript{WDQL}, or on all existing datasets and our dataset as GRISP\textsubscript{All}. In terms of F\textsubscript{1}-score, GRISP\textsubscript{All} shows a strong performance across the board, and does not massively fall behind one of the other models on any dataset. GRISP\textsubscript{WDQL} clearly outperforms GRISP\textsubscript{Prev} on QAWiki, SPINACH, and WDQL, and even QALD-7 and QALD-10, even though it was not trained on the latter two.

Surprisingly, we find that GRISP\textsubscript{Prev} achieves much higher F\textsubscript{1}-scores than GRISP\textsubscript{WDQL} on SimpleQuestions, LC-QuAD 2.0, and WWQ, although these datasets seem to be easier than the other ones judging from the statistics (lower share of advanced queries, see \Cref{tab:stats}, and of distinct query patterns, see \Cref{tab:ext}).
Below, we provide evidence that this is not because GRISP\textsubscript{WDQL} generates bad SPARQL queries for these datasets, but because these three datasets expect SPARQL queries following dataset-specific conventions or patterns that are different from natural query formulations and can only be learned from the respective training sets.
For example, \cite{modernsempar} showed that LC-QuAD 2.0 queries can be generated with 92\% F\textsubscript{1}, when gold entities and properties are given, suggesting that query generation is actually easy on this benchmark and
follows repetitive patterns that can be learned well from seeing similar training data. Consequently, if conventions are not followed, it is easy to get low F\textsubscript{1}-scores despite producing sensible SPARQL queries. This also explains why GRISP\textsubscript{All} almost closes the gap to GRISP\textsubscript{Prev} on these three datasets.

To verify this, we perform an LLM-as-a-judge evaluation between GRISP\textsubscript{Prev} and GRISP\textsubscript{WDQL} across all datasets. 
We provide GPT-4.1 mini with the question, both generated SPARQL queries (with IRI descriptions and execution
results), but without the ground truth SPARQL query, and ask it to judge which model's output is better or whether
both are equally good.
In this evaluation, the SPARQL queries generated by GRISP\textsubscript{WDQL} are preferred on average over the ones by GRISP\textsubscript{Prev} across all datasets, including SimpleQuestions, WWQ, and LC-QuAD 2.0. This suggests that these three datasets rely more on learning expected SPARQL query patterns than on natural generalization capabilities. It also aligns with the characterization from \cite{kgqadatasets} of SimpleQuestions and WebQuestionsSP (the basis for WWQ) as only testing IID generalization, where the test set follows the training distribution and all relevant SPARQL constructs, classes, and properties have already been seen during training.

Finally, we investigate how the training set size impacts performance when training a GRISP model on WDQL. We find that training on a larger subset generally yields better performance. For WDQL the F\textsubscript{1}-score strictly increases with training set size, while for the other datasets the correlation is weaker. Similar to the effects discussed above, a lower F\textsubscript{1}-score for a larger training subset does not necessarily mean worse SPARQL generation quality. For example, on QAWiki we find the outputs of the model trained on 80k pairs to be preferred over those of the model trained on 40k pairs in an LLM-as-a-judge evaluation (23.6\% vs. 17.0\%, 59.4\% ties) despite the lower F\textsubscript{1}-score. Note that training on as few as 10k WDQL pairs already yields a model that performs better on QALD-7, QALD-10, QAWiki, and SPINACH than GRISP\textsubscript{Prev} from \Cref{tab:qa-combined}, which was trained on all previous Wikidata datasets combined. See the details in \Cref{tab:subset}.

The GRISP code is available as part of the GRASP repository at \grasp. We also provide the evaluations and outputs of GRISP at \graspeval, a demo of GRISP\textsubscript{WDQL} at \grisp{}, and all model checkpoints at \grispdata{}.

\begin{table}
	\caption{F\textsubscript{1}-score of GRISP trained on random subsets of the WDQL training set of increasing size.}
	\begin{threeparttable}
\begin{tabular}{lccccc}
\toprule
\textbf{Size} & \textbf{QALD-7} & \textbf{QALD-10} & \textbf{QAWiki} & \textbf{SPIN.} & \textbf{WDQL}\tnote{1} \\
\midrule
10k & 65.5 & 45.0 & 39.0 & 30.0 & \cellcolor{gray!15}51.6 \\
20k & 65.4 & 46.4 & 42.9 & \textbf{32.4} & \cellcolor{gray!15}53.5 \\
40k & \underline{68.1} & \textbf{50.5} & \textbf{48.4} & \underline{32.0} & \cellcolor{gray!15}\underline{55.5} \\
80k & \textbf{70.1} & \underline{49.9} & \underline{47.3} & 29.4 & \cellcolor{gray!15}\textbf{57.6} \\
\end{tabular}
\end{threeparttable}
\vspace{0.1cm}
\par
\centering
\begin{minipage}[t]{\columnwidth}\vspace{0pt}
\footnotesize\raggedright
Bold and underlined values indicate the highest and second-highest scores for each dataset, respectively. Grey cells represent in-distribution evaluation.\\
\textsuperscript{1}~Evaluated on 1000 randomly selected test samples due to the large test set size.
\end{minipage}

	\label{tab:subset}
\end{table}

\section{Conclusion}

This work presents WDQL, the largest dataset of question-query pairs for Wikidata to date with over 335k pairs. We show that question-answering methods can be improved significantly when training on WDQL. To build the dataset, we develop an S2Q agent capable of generating natural-language questions given a (potentially anonymized) SPARQL query. We believe that WDQL will benefit future research on question answering and beyond. Since our approach is agnostic to the underlying dataset, we would also like to see it applied to datasets other than Wikidata in future work.

\section*{Statement on the use of AI}
We did not use AI for core parts of the work. In particular, we did not use AI to implement the question-query pair generation in \grasp, and not for writing. We did use AI coding assistants in an interactive manner to help us with writing scripts for data processing and visualization. This includes most scripts in \github. All code was carefully checked for quality and correctness, and all critical parts (such as the calculation of dataset statistics) were verified via unit tests.

\begin{acks}
Funded by the Deutsche Forschungsgemeinschaft (DFG, German Research Foundation) – Project-ID 499552394 – SFB 1597.
\end{acks}

\bibliographystyle{ACM-Reference-Format}
\bibliography{wdql}

\end{document}